\documentclass[runningheads]{llncs}

\usepackage{eccv}

\usepackage{eccvabbrv}

\usepackage{graphicx}
\usepackage{colortbl}
\usepackage{algorithm}
\usepackage{algpseudocode}
\usepackage{booktabs}
\usepackage{makecell}
\usepackage{booktabs}
\usepackage{subcaption}
\usepackage[table]{xcolor} 
\usepackage{graphicx} 
\usepackage{bm}

\usepackage[accsupp]{axessibility}  

\usepackage{hyperref}

\usepackage{orcidlink}

\begin{document}


\title{UPGS: Unified Pose‑aware Gaussian Splatting for Dynamic Scene Deblurring}
\titlerunning{Abbreviated paper title}

\author{
Zhijing Wu\inst{1} \and
Longguang Wang\inst{2}
}

\authorrunning{Z. Wu and L. Wang}

\institute{
University of Cambridge, Cambridge, UK\\
\email{zw436@cantab.ac.uk}
\and
The Shenzhen Campus of Sun Yat-sen University, Shenzhen, China\\
\email{wanglg9@mail.sysu.edu.cn}
}

\maketitle

\begin{abstract}
      Reconstructing dynamic 3D scenes from monocular video has broad applications in AR/VR, robotics, and autonomous navigation, but often fails due to severe motion blur caused by camera and object motion. Existing methods commonly follow a two‑step pipeline, where camera poses are first estimated and then 3D Gaussians are optimized. Since blurring artifacts usually undermine pose estimation, pose errors could be accumulated to produce inferior reconstruction results. To address this issue, we introduce a unified optimization framework by incorporating camera poses as learnable parameters complementary to 3DGS attributes for end-to-end optimization. Specifically, we recast camera and object motion as per‑primitive SE(3) affine transformations on 3D Gaussians and formulate a unified optimization objective. For stable optimization, we introduce a three‑stage training schedule that optimizes camera poses and Gaussians alternatively. Particularly, 3D Gaussians are first trained with poses being fixed, and then poses are optimized with 3D Gaussians being untouched. Finally, all learnable parameters are optimized together. Extensive experiments on the Stereo Blur and BARD-GS dataset with challenging real‑world sequences demonstrate that our method achieves significant gains in reconstruction quality and pose estimation accuracy over prior dynamic deblurring methods.
\end{abstract}

\section{Introduction}
\label{sec:intro}

Recent advances in explicit 3D Gaussian Splatting (3DGS) \cite{kerbl20233d} and implicit Neural Radiance Fields (NeRF) \cite{mildenhall2021nerf} have pushed photorealistic reconstruction from static 3D scenes to fully dynamic 4D ones. These techniques promote emerging applications such as mobile AR/VR, autonomous driving, and robotics, in which data are almost acquired with a single, moving camera in unconstrained environments \cite{matsuki2024gaussian, jiang2024vr, zhou2024drivinggaussian}. In these scenarios, severe motion-induced blur usually remains a dominant factor that limits reconstruction quality.

Motion blur arises as the sensor integrates light over a finite exposure while the object or the camera moves, which consists of two components: camera-induced blur from ego-motion and object-induced blur from fast, often non-rigid motion \cite{ji2024motion, rozumnyi2022motion}. These motions produce blurring artifacts that hinder high-quality 4D reconstruction. The major reasons are twofold: \textbf{(i) Low pose accuracy}. Structure-from-motion pipelines such as COLMAP depend on sharp, repeatable keypoints to triangulate a sparse cloud and estimate camera poses. However, motion blur loses high-frequency details, frustrates feature matching, and produces inaccurate camera poses \cite{peng2023pdrf, lee2024sharp}. \textbf{(ii) High ill-posedness}. As each blurred pixel integrates light from a trajectory rather than a fixed 3D location, which weakens the geometric constraint and increases the ill-posedness to reconstruct 4D scenes \cite{kumar2025dynamode}.

Over the last years, numerous efforts \cite{ma2022deblur, lee2023exblurf, lee2023dp, wang2024mp, wang2023bad, lee2024deblurring, oh2024deblurgs} have been made to reconstruct sharp scenes from motion-blurred imagery, yet most methods still focus on static 3D scenes. Recently, dynamic deblurring has emerged \cite{luthra2024deblur, kumar2025dynamode, sun2024dyblurf, bui2025moblurf, luo2024dynamic} and is mainly built on NeRF. Since the continuous representation naturally favors smooth, low-frequency results, fast object motions are typically more difficult to be optimized, leading to unsatisfactory solutions. In addition, volumetric ray marching through a deep MLP requires long training time \cite{mildenhall2021nerf}, making deployment for time-sensitive AR/VR use cases challenging. Recent advances in Gaussian splatting has provided a new approach, yet their applications in dynamic deblurring is still under-investigated.

Intuitively, 4D reconstruction from a blurry video is a joint optimization problem in which camera pose estimation and sharp scene reconstruction are deeply coupled \cite{lin2021barf, bian2023nope}. On the one hand, sharp reconstruction improves the accuracy of pose estimation by providing higher-quality keypoints. On the other hand, accurate camera poses alleviate the ill-posedness to facilitate sharp images to be reconstructed. However, most prior approaches ignore the interaction between pose estimation and scene reconstruction. These methods commonly employ COLMAP poses to reconstruct the scene without further pose refinement. As a result, pose errors can be accumulated to degrade the final reconstruction results.

In this paper, we present \textbf{UPGS}, a Unified Pose-aware Gaussian Splatting framework for dynamic scene deblurring. Specifically, to avoid error accumulation caused by erroneous COLMAP poses, we formulate camera and object motions as learnable per-primitive SE(3) affine transformations on 3D Gaussians and incorporate them for joint optimization.

As affine transformations are highly coupled with 3D Gaussians, an alternative training scheme is employed. Particularly, we first leave out affine transformations to optimize 3D Gaussians using initial COLMAP poses. In this way, static regions can be roughly reconstructed while dynamic regions still suffer poor quality. Then, we include time-conditioned SE(3) transformations for training while fixing 3D Gaussians, refining the geometry of dynamic regions. Finally, both affine transformations and 3D Gaussians are jointly optimized. Extensive evaluations on the Stereo Blur benchmark and challenging real-world sequences show that our UPGS achieves superior reconstruction fidelity and pose accuracy compared to existing dynamic deblurring methods.

\begin{itemize}
  \item We propose \textbf{UPGS}, a Unified Pose‑aware Gaussian Splatting framework that jointly optimizes camera poses and 3D Gaussian scene parameters in an end‑to‑end manner to reconstruct dynamic scenes from blurry monocular video.
  \item We recast both camera and object motion as per‑primitive $\mathrm{SE}(3)$ affine warps on 3D Gaussians, enabling joint refinement in the same optimization pipeline.
  \item We introduce a three‑stage training schedule—scene‑only, pose‑only, then full joint optimization—to prevent error accumulation and accelerate convergence.
  \item We demonstrate that UPGS yields sharp reconstructions and more accurate camera‑pose estimates on challenging real‑world sequences from the Stereo Blur \cite{zhou2019davanet} and BARD‑GS datasets \cite{lu2025bard}, indicating its effectiveness for dynamic deblurring tasks.
\end{itemize}

\section{Related Work}

In this section, we first review previous 2D deblurring methods, and then present advanced novel view synthesis methods. Finally, we focus on discussing recent advances regarding deblurring dynamic reconstruction.

\subsection{2D Deblurring}

The goal of deblurring is to recover sharp imagery from blur caused by camera shake or object motion. Classical methods cast this as an inverse‑filtering problem with hand‑crafted priors \cite{fergus2006removing, krizhevsky2012advances, pan2016blind, michaeli2014blind}, while modern deep‑learning approaches train end‑to‑end multi‑scale, recurrent, attention‑ and adversarial‑based networks \cite{nah2017deep, tao2018scale, zamir2021multi, kupyn2018deblurgan} on large datasets (e.g GoPro \cite{nah2017deep}, DeblurGAN \cite{kupyn2018deblurgan}), enabling real‑time single‑image deblurring. Video deblurring further leverages temporal cues by aligning or fusing features across frames via optical flow, deformable convolutions, or recurrent/3D networks \cite{su2017deep, wang2019edvr, pan2020cascaded, zhou2019spatio, nah2019recurrent}. Despite these advances in 2D image/video space, such models lack explicit 3D scene geometry modeling. Our work formulates deblurring as a 4D reconstruction problem within the Gaussian Splatting framework.

\subsection{Dynamic Novel View Synthesis}

Gaussian Splatting (GS) \cite{kerbl20233d} has emerged as a popular choice for modeling 3D scenes due to its explicit structure, high rendering efficiency, and real-time performance. Recent works have extended GS into 4D to handle dynamic scene reconstruction from monocular input. Early attempts, such as Dynamic 3D Gaussians \cite{luiten2024dynamic}, independently tracked each Gaussian across timestamps to capture scene motion, while Deformable-3D-Gaussians \cite{yang2024deformable} introduced a deformation field via an MLP to learn time-dependent transformations of canonical Gaussians. 4D Gaussian Splatting \cite{wu20244d, yang2023real, wang2024shape} advanced this idea by directly lifting Gaussians into 4D space, jointly encoding spatial and temporal variations to enable high-quality real-time dynamic rendering. More recent developments, including Disentangled 4D Gaussian Splatting \cite{feng2025disentangled}, significantly improve efficiency by factorizing spatial and temporal components, while benchmarking efforts \cite{liang2024monocular} have analyzed the performance and brittleness of various monocular 4D GS pipelines. These milestones have collectively established GS as a competitive framework for 4D reconstruction, capable of modeling complex motions while maintaining real-time rendering capabilities. However, existing 4D GS approaches still rely on sharp, high-quality inputs and accurate camera poses to achieve good reconstruction fidelity, which limits their robustness in real-world scenarios involving motion blur. To address this, our method aims to reconstruct 4D scenes directly from blurry monocular inputs, jointly handling deblurring and dynamic scene modeling.

\subsection{Deblurring Dynamic Reconstruction}

Recently, a variety of works have been proposed to recover high-quality 3D reconstructions from blurred inputs. NeRF-based approaches \cite{ma2022deblur, wang2023bad, lee2023dp, lee2023exblurf} treat blur as part of the forward model by learning spatially varying kernels or simulating exposure via camera-trajectory subframes. 3DGS-based static deblurring folds the same exposure model into an explicit pipeline, with Debluring 3DGS \cite{lee2024deblurring} as a pioneering example. BAD-Gaussians \cite{zhao2024bad} and DeblurGS \cite{oh2024deblurgs} share a workflow that jointly estimates exposure-time camera motion and optimizes Gaussian scene parameters by rendering virtual sharp subframes along a continuous SE(3) trajectory, averaging them to synthesize the observed blur, and driving a photometric loss against the input images. These approaches deliver high-quality results on static scenes but do not handle dynamics. Turning to dynamic 4D deblurring, \cite{sun2024dyblurf, bui2025moblurf, luthra2024deblur} generates latent rays within the exposure and performs time-aware volume rendering. However, training often requires significantly longer runtimes, and the implicit MLP’s smoothness and capacity bias toward low-frequency content causes fast non-rigid motions to be underfit. Our approach fills this gap with a 4D Gaussian-splatting formulation that reconstructs directly from blurry monocular video while explicitly decoupling camera-induced blur via subframe warps along with the object motion, improving efficiency and preserving high-frequency structure in dynamic areas.

\begin{figure*}[t]
  \centering
  \includegraphics[width=1.01\linewidth,height=0.4\textheight]{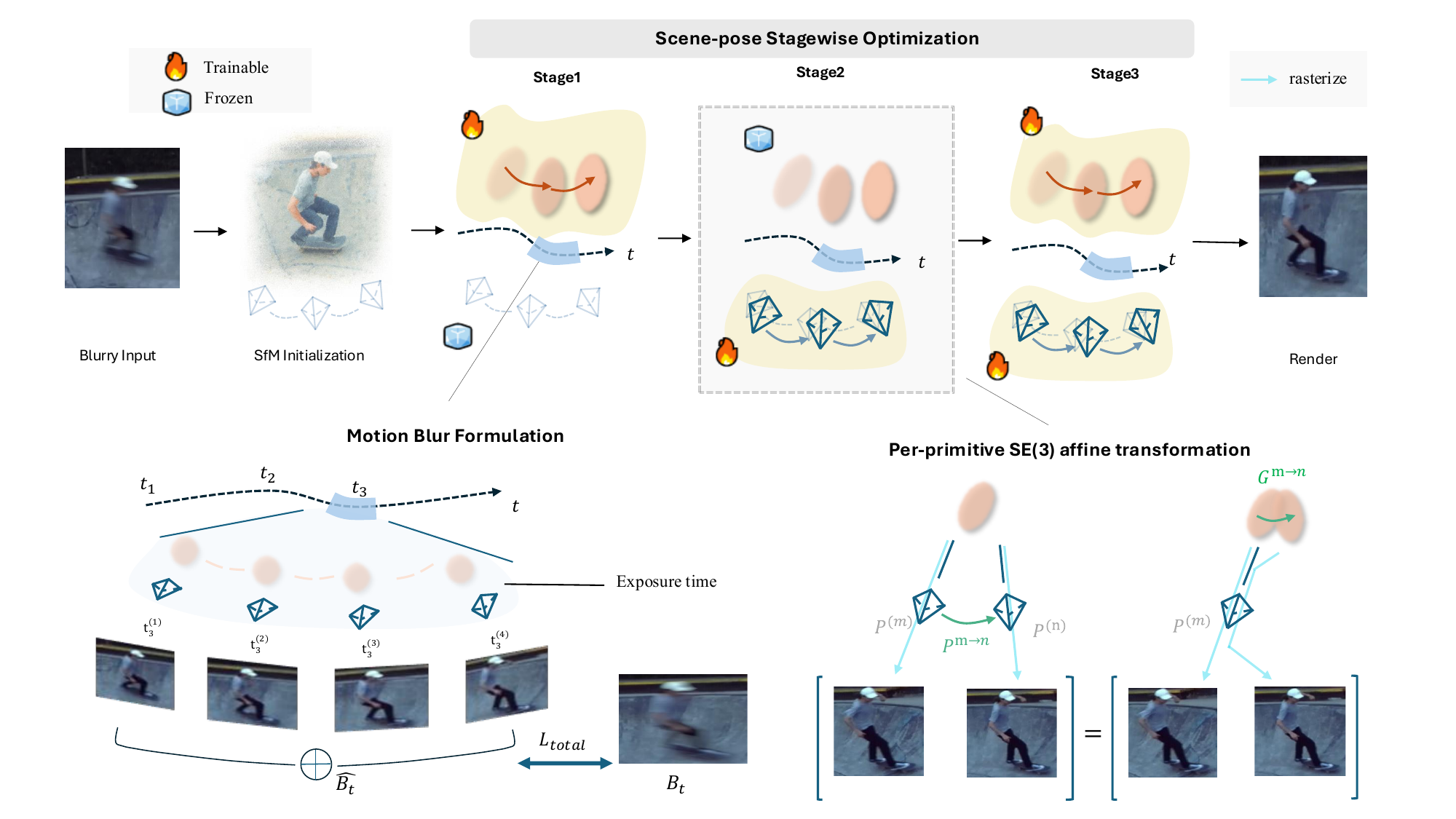} 
  \caption{\textbf{An overview of UPGS.} We adopt a three-stage training schedule for optimization. Camera motion is represented as trainable SE(3) affine transformations (Sec 3.2) on Gaussian primitives, thereby camera poses can be optimized together with the reconstructed scene. Using COLMAP poses for initialization, we first optimize Gaussian primitives with poses fixed. Next, with Gaussian primitives being frozen, we refine only the affine transformations. In the final stage, we jointly fine-tune scene and pose so they co-adapt (Sec 3.3), yielding sharper renders, higher reconstruction fidelity, and more accurate camera trajectories.}
  \label{fig:overview}
\end{figure*}

\section{Method}

In this section, we first present the preliminaries in Sec. 3.1. Then, we introduce our unified pose-aware framework in Sec. 3.2 and detail our optimization strategy is Sec. 3.3.

\subsection{Preliminaries}

\subsubsection{Gaussian Splatting Primer}
3D Gaussian Splatting (3DGS) encodes a scene as a collection of anisotropic \(3\text{D}\) Gaussians.
Each primitive is parameterised by a centre \(\mathbf{x}\in {R}^{3}\), covariance \(\boldsymbol{\Sigma}\in{R}^{3\times3}\), opacity \(\alpha\), and view-dependent colour using learnable spherical-harmonic (SH) coefficients.
Under a given camera pose, every Gaussian is projected to the image plane, yielding a \(2\text{D}\) footprint whose shape is controlled by the projected covariance \(\boldsymbol{\Sigma}'\).
For a pixel location \(\mathbf{x}\), colour is accumulated with front-to-back alpha blending:
\begin{equation}
  \hat{c}(\mathbf{x}) \;=\;
  \sum_{i\in\mathcal{N}} c_{i}\,\alpha_{i}(\mathbf{x})
  \prod_{j=1}^{\,i-1}\!\bigl(1-\alpha_{j}(\mathbf{x})\bigr),
  \label{eq:alpha_blend}
\end{equation}
where \(c_i\) is the SH-predicted colour and \(\alpha_i(\mathbf{x})\) is the opacity of the \(i\)-th Gaussian.
Scene parameters \(\{\mathbf{x},\boldsymbol{\Sigma},\alpha,\text{SH}\}\) are optimised by minimising an image-space loss between the rasterized images and their reference counterparts across all viewpoints.

\begin{algorithm}[t]
\caption{Stagewise Pose--Scene Optimization}
\label{alg:concise_stagewise}
\begin{algorithmic}[1]
\Require scene $\theta$, poses $\phi$, epochs $E_{\max}$, stop-gradient operator $\bot(\cdot)$
\State $\text{stage}\gets 1$
\For{$e=0$ \textbf{to} $E_{\max}-1$}
  \For{$t$ \textbf{in minibatch}}
    \If{$\text{stage}=1$} \Comment{scene warm-up}
      \State $\theta \leftarrow \mathcal{U}_\theta\!\big(L_{\text{total}}(t;\theta,\bot(\phi))\big)$
      \If{$\Delta_{\text{psnr}}<\epsilon_1 \wedge \text{anchor}>\tau_a$} \State $\text{stage}\gets 2$ \EndIf
    \ElsIf{$\text{stage}=2$} \Comment{pose refinement}
      \State $\phi \leftarrow \mathcal{U}_\phi\!\big(L_{\text{total}}(t;\bot(\theta),\phi)\big)$
      \If{$\Delta_{\text{pose}}<\epsilon_2 \wedge \mathcal{P}[\mathcal{L}]>\tau_{\text{gain}}$} \State $\text{stage}\gets 3$ \EndIf
    \Else \Comment{joint fine-tune}
      \State $\theta \leftarrow \mathcal{U}_\theta\!\big(L_{\text{total}}(t;\theta,\phi)\big)$; \;
             $\phi \leftarrow \mathcal{U}_\phi\!\big(L_{\text{total}}(t;\theta,\phi)\big)$
    \EndIf
  \EndFor
\EndFor
\end{algorithmic}
\end{algorithm}

\subsubsection{Gaussian Deformation for Object Motion}
Following Deblur4DGS \cite{wu2024deblur4dgs}, object deformation is modeled by \(\mathrm{SE}(3)\) transforms \(\{A_t, E_t\}\) via Shape‐of‐Motion \cite{wang2024shape}, i.e.
\begin{equation}
  \bm{G_{\mathrm{dym},t}} = A_t\,\bm{G_{\mathrm{dym},c}} + E_t.
  \label{eq:gaussian_deform}
\end{equation}
Here $A_t \in \mathbb{R}^{3\times 3}$ is the rigid rotation matrix and $E_t \in \mathbb{R}^{3}$ is the translation vector at time $t$. The canonical dynamic Gaussians \(\bm{G_{\mathrm{dym},c}}\) are obtained by uniformly dividing the video into \(L\) segments and selecting the frame with the highest Laplacian sharpness in each segment as the reference. Rather than learning a separate \(\mathrm{SE}(3)\) for every subframe, we estimate only the start‑ and end‑of‑exposure parameters (with weights \(\pm w_t/2\)) and interpolate intermediate subframe deformations by
\begin{equation}
  w_{t,i}
  = \Bigl(1 - \frac{i-1}{N-1}\Bigr)\odot\frac{w_t}{2}
  \;+\;
  \frac{i-1}{N-1}\odot\Bigl(-\frac{w_t}{2}\Bigr).
  \label{eq:interp_weights}
\end{equation}
Canonical Gaussians, \(\mathrm{SE}(3)\) deformation modules, and exposure weights \(w_t\) are then optimized jointly using dynamic reconstruction loss.

\subsubsection{Motion Blur Formation}

Motion blur arises because, for a frame stamped at time \(t\), the sensor integrates light over an exposure interval while the camera or scene moves. The exposure duration at time \(t\) is denoted as \(\delta_t\).
Let \(I_t(u,v)\) be the instantaneous rendering at pixel \((u,v)\) at time \(t\).
The recorded blurry pixel is the temporal integral
\begin{equation}
  B_t(u,v) \;=\; \phi \int_{0}^{\delta_t} I_t(u,v)\,d\delta,
  \label{eq:blur_integral}
\end{equation}
with normalisation factor \(\phi\).
In practice we discretise the exposure time \(\delta_t\) at global time t into \(N\) latent sharp frames \(\{I_i\}_{i=0}^{N-1}\) and approximate
\begin{equation}
  B_t(u,v) \;\approx\; \frac{1}{N}\sum_{i=0}^{N-1} I_n(u,v).
  \label{eq:blur_discrete}
\end{equation}
This discrete form enables differentiable simulation of motion blur: sharp images rendered at intermediate timestamps are averaged to reproduce the observed blurry input.

\subsection{Unified Pose-Aware Framework}
\label{sec:Unified}

Our unified pose-aware framework is illustrated in Fig.~\ref{fig:overview}. To make camera pose trainable within the 3DGS framework, we formulate the camera motion as an affine transformation of the scene under a given fixed pose. In this way, camera motion can be modeled with object motion using unified representations, which are then applied to Gaussian primitives. Given a blurry frame \(B_t\) at timestamp $t$, we estimate the camera pose using CLOMAP and employ it to initialize the affine transformation \(P_t \in \mathrm{SE}(3)\). To capture the camera trajectory during exposure time, we subdivide the exposure time into \(N\) intervals such that the camera poses for subframes can be denoted as:
\begin{equation}
  P_t^{(m)} = \bigl[R_t^{(m)},\,t_t^{(m)}\bigr] \quad \text{for } m=1,\dots,N,
  \label{eq:subframe_poses}
\end{equation}
Without loss of generality, one of these poses is selected as a reference pose:
\begin{equation}
  P_t^{(n)} = \bigl[R_t^{(n)},\,t_t^{(n)}\bigr].
  \label{eq:reference_pose}
\end{equation}
Crucially, for any two poses \(P^{(n)}\) and \(P^{(m)}\), there exists an affine transformation \(T^{n\to m}\) on the Gaussians \(\bm{G} = \{\bm{G}_{\mathrm{sta}}, \bm{G}_{\mathrm{dym}}\}\) to satisfy:
\begin{equation}
  I\bigl(\bm{G},\,P^{(m)}\bigr)
  =
  I\!\bigl(T^{n\to m}(\bm{G}),\,P^{(n)}\bigr),
  \label{eq:equivalent_render}
\end{equation}
where $I\bigl(\bm{G},\,P^{(m)}\bigr)$ denotes the image rasterization using 3DGS $\bm{G}$ under the pose $P^{(m)}$. Specifically, we employ the affine transformation to map the reference view into the \(m\)-th subframe view,
\begin{equation}
  \begin{aligned}
    R_t^{\,n\to m} & = \bigl(R_t^{(m)}\bigr)^{\!\top}\,R_t^{(n)},                      \\
    t_t^{\,n\to m} & = \bigl(R_t^{(m)}\bigr)^{\!\top}\bigl(t_t^{(m)} - t_t^{(n)}\bigr)
  \end{aligned}
  \label{eq:affine_rt_tt}
\end{equation}

\begin{figure*}[t]
  \centering
  \includegraphics[width=1\linewidth,height=0.5\textheight,keepaspectratio]{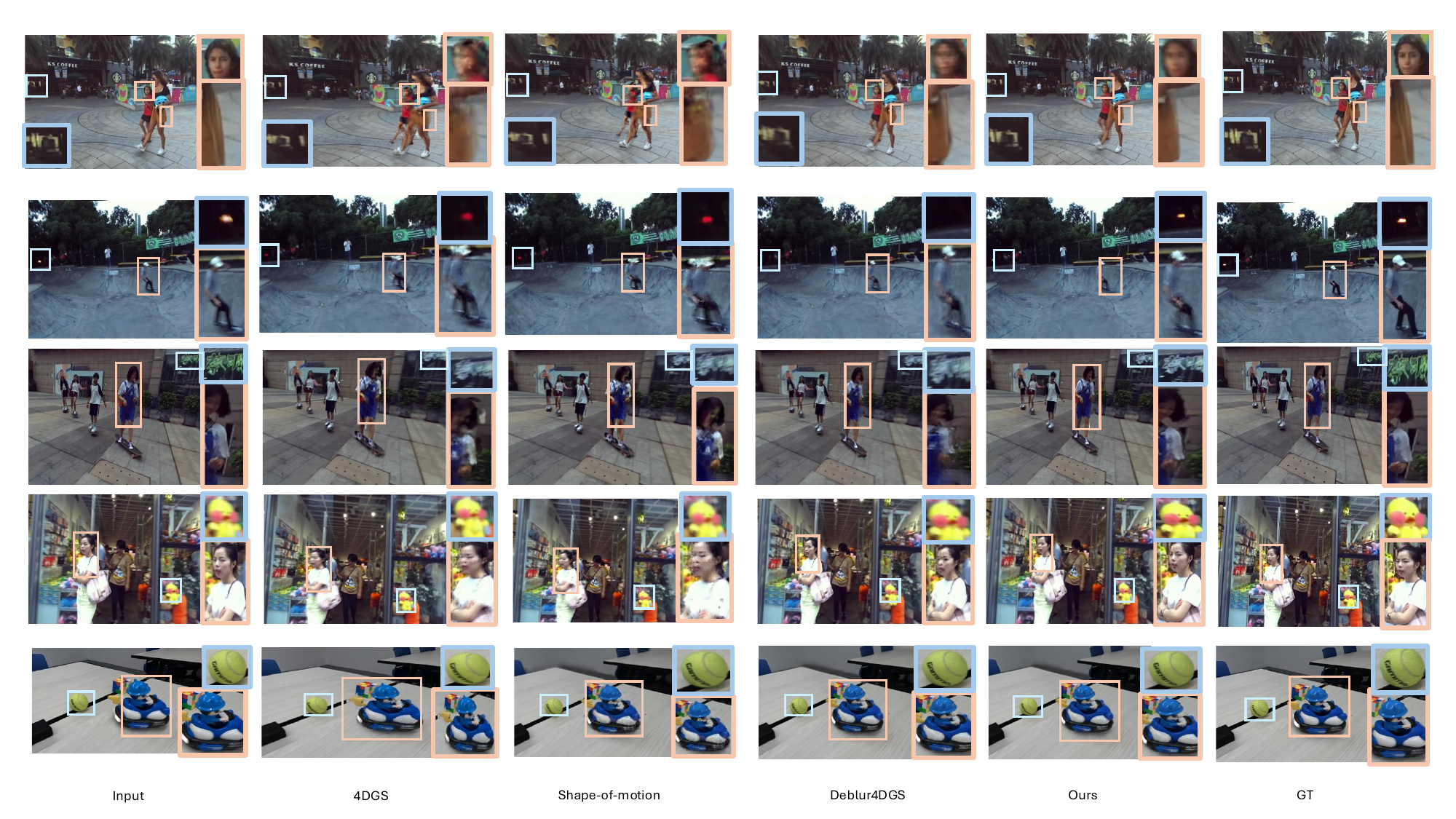} 
  \caption{\textbf{Visual Comparison on Stereo Blur and BARD-GS Dataset}. The \colorbox{orange!30}{orange} boxes highlights regions with intense dynamic motion and the \colorbox{blue!15}{blue} boxes indicate purely static areas.}
  \label{fig:sota}
\end{figure*}

and apply these transformations directly to each Gaussian primitive \(\bm{G}=(\mu_i,\Sigma_i,c_i,\alpha_i)\) via
\begin{equation}
  \begin{aligned}
    \mu_i'    & = R_t^{\,n\to m}\,\mu_i + t_t^{\,n\to m},                       \\
    \Sigma_i' & = R_t^{\,n\to m}\,\Sigma_i\,\bigl(R_t^{\,n\to m}\bigr)^{\!\top}
  \end{aligned}
  \label{eq:affine_mu_sigma}
\end{equation}
leaving color \(c_i\) and opacity \(\alpha_i\) unchanged. As a result, by rendering the warped 3D Gaussians \(\{\bm{G}'\}\)  from the fixed reference pose \(P_t^{(n)}\) and averaging over all intermediate subframes, we faithfully mimic the true motion blur—\ the integral of the scene’s geometry under the camera’s continuous motion—directly in the geometry domain.

\subsection{Optimization Strategy}
\label{sec:stagewise}

Since pose and geometry are tightly coupled, a fully joint optimization can be  highly ill‑posed and produce trivial solutions. Minor errors in camera pose estimation may skew the scene fit, which in turn misguides subsequent pose updates, leading to error accumulation and unsatisfactory reconstruction results. To address this, we partition the training phase into three successive stages: Stage 1 optimizes only the scene parameters, warm up the model with the basic scene geometry. Stage 2 freezes the scene and updates only the camera poses, enabling the system to digest the camera motion and understand the source of blur. Stage 3 then performs full joint optimization of both scene and pose. The overall training strategy is summarized in Algorithm \ref{alg:concise_stagewise}.

The transition between stages is determined automatically by two adaptive gates. Gate 1 monitors whether the scene representation has saturated. It checks: 

\begin{itemize}
  \item $\Delta_{\mathrm{psnr}} < \epsilon_1$: whether the increase in validation PSNR has plateaued.
  \item $\mathrm{anchor} > \tau_a$: the extent to which the static Gaussians begin to shift as the model attempts to explain residual errors by implicitly mimicking camera motion.
\end{itemize}

\vspace{-10pt}
{\setlength{\abovedisplayskip}{8pt}
 \setlength{\belowdisplayskip}{8pt}
\begin{equation}
\begin{aligned}
\text{anchor}
=
\sum_{g \in G_{\text{static}}}
\left( \lVert \Delta \mu_g \rVert^{2}
     + \lVert \Delta R_g \rVert^{2} \right)
\end{aligned}
\end{equation}
}

\vspace{20pt}

\noindent Once these criteria are met, the scene has extracted the majority of learnable appearance structure, and pose optimization is activated.

Gate 2 evaluates whether pose refinement is exhausted. It detects: 

\begin{itemize}
  \item $\Delta_{\mathrm{pose}} < \epsilon_2$: whether the decrease in pose error has plateaued.
  \item $\mathcal{P}[\mathcal{L}](\theta,\phi)> \tau_{\text{gain}}$: whether enabling scene updates would yield a meaningful improvement in the overall objective. To assess this, we simulate a virtual optimizer step on the scene parameters as if scene learning were enabled. We then evaluate the corresponding decrease in loss by computing a lightweight look-ahead gain proxy from the scene gradients. A marked improvement is identified when incorporating scene parameters results in a clear reduction in total loss.
\end{itemize}

\vspace{-15pt} 
\begin{equation}
\begin{aligned}
\mathcal{P}[\mathcal{L}](\theta,\phi)
= L_{\mathrm{pose}}(\phi)
- L_{\mathrm{joint}}(\theta,\phi)
\end{aligned}
\end{equation}

\noindent When both conditions hold, the optimizer proceeds to Stage 3 for joint refinement of geometry and camera motion. The detailed values for the gating hyperparameters are provided in the supplementary material.

\begin{table}[t] 
  \centering
  \caption{Quantitative results on the Stereo and BARD datasets. \colorbox{red!20}{Red} and \colorbox{yellow!20}{yellow} denote the best and second-best per column, respectively.}
  \label{tab:avg_metrics}
  \resizebox{\textwidth}{!}{%
    \setlength{\tabcolsep}{3.5pt}%
    \begin{tabular}{@{}|l|cccc|cccc|@{}}
      \hline
      & \multicolumn{4}{c|}{\textbf{Stereo Blur Dataset}}
      & \multicolumn{4}{c|}{\textbf{BARD Dataset}} \\
      \cline{2-9}
      & \textbf{PSNR} $\uparrow$
      & \textbf{SSIM} $\uparrow$
      & \textbf{LPIPS} $\downarrow$
      & \textbf{\shortstack{Training \\ time}} $\downarrow$
      & \textbf{PSNR} $\uparrow$
      & \textbf{SSIM} $\uparrow$
      & \textbf{LPIPS} $\downarrow$
      & \textbf{\shortstack{Training \\ time}} $\downarrow$ \\
      \hline
      DyBluRF         & 23.82 & 0.690 & 0.471 & 49
                      & 16.25 & 0.523 & 0.971 & 87 \\
      Shape-of-Motion & 27.49 & \cellcolor{red!20}0.922 & 0.192 & \cellcolor{yellow!20}3.7
                      & 17.92 & 0.718 & \cellcolor{yellow!20}0.434 & \cellcolor{yellow!20}8 \\
      4DGS            & 26.33 & 0.733 & 0.345 & \cellcolor{red!20}0.8
                      & 17.28 & 0.609 & 0.877 & \cellcolor{red!20}1.8 \\
      Deblur4DGS      & \cellcolor{yellow!20}29.02 & 0.745 & \cellcolor{yellow!20}0.165 & 6.9
                      & \cellcolor{yellow!20}18.80 & \cellcolor{yellow!20}0.720 & \cellcolor{red!20}0.372 & 13 \\
      \textbf{Ours}   & \cellcolor{red!20}\textbf{30.14}
                      & \cellcolor{yellow!20}\textbf{0.911}
                      & \cellcolor{red!20}\textbf{0.107}
                      & 4.2
                      & \cellcolor{red!20}19.50
                      & \cellcolor{red!20}0.731
                      & 0.550
                      & 10 \\
      \hline
    \end{tabular}%
  }
\end{table}

\vspace{0.6em}
To avoid over‑smoothed edges between the static and dynamic regions of the scene, we train background and foreground Gaussians in an end‑to‑end pipeline. Particularly, we learn \(G_{\mathrm{sta}}\) and \(G_{\mathrm{dym}}\) jointly by minimizing

\begin{equation} 
\begin{aligned}
&L_{\mathrm{dym}}\bigl(\bm{G_{\mathrm{sta}}},\,\bm{G_{\mathrm{dym}}}; P_t^{(m)}\bigr) 
\\
= &\bigl\lVert I\bigl(\bm{G_{\mathrm{sta}}},\,\bm{G_{\mathrm{dym}}}; P_t^{(m)}\bigr) - I(t) \bigr\rVert^{2},
\end{aligned}
\label{eq:loss_dym} 
\end{equation}

\noindent which regularizes all motion‑induced blur over sub‑frames \(m\). To further enforce a crisp background, we introduce
\begin{equation}
  L_{\mathrm{static}}\bigl(\bm{G_{\mathrm{sta}}}; P_t\bigr)
  = \bigl\lVert I\bigl(\bm{G_{\mathrm{sta}}}; P_t\bigr)
  - I_{\mathrm{static}}(t)\bigr\rVert^2,
  \label{eq:loss_static}
\end{equation}
The final objective is defined as:
\begin{equation}
  L_{\mathrm{total}}
  = L_{\mathrm{dym}}\bigl(\bm{G_{\mathrm{sta}}},\,\bm{G_{\mathrm{dym}}}\bigr)
  + L_{\mathrm{static}}\bigl(\bm{G_{\mathrm{sta}}}\bigr),
  \label{eq:loss_total}
\end{equation}
which is optimized end‑to‑end so that background and foreground Gaussians co‑adapt, capturing dynamic blur while preserving static details.

\subsection{Discussion}

We position UPGS against existing joint scene-pose deblurring pipelines by highlighting how it departs from prior neural-based approaches. BARF \cite{lin2021barf} resolves pose–scene entanglement by gradually unmasking high-frequency positional encodings inside NeRF. However, it applies only to sharp, static input. We perform joint optimisation and reconstruction from dynamic, motion-blurred video. Unlike Deblur4DGS \cite{wu2024deblur4dgs} which first optimises a frozen static background and then deblurs the dynamic foreground with extra regularisation, UPGS trains static and dynamic Gaussians together from the outset, avoiding additional loss terms. The recent BARD-GS \cite{lu2025bard} learns one global pose per frame in camera space and interpolates intermediate viewpoints, while we represents both scene and pose in a unified geometry space and directly captures the entire camera trajectory.

\section{Experiment}

In this section, we first introduce the implementation details, and then compare the performance of our method against previous approaches. Finally, we conduct ablation experiments to demonstrate the effectiveness of our method designs.

\subsection{Implementation details}

\textbf{Dataset.} We evaluate our method on the Stereo Blur dataset \cite{sun2024dyblurf} and BARD-GS dataset \cite{lu2025bard}. Stereo Blur dataset comprises six scenes exhibiting significant motion blur from both camera and object motion. Each sequence was captured with a ZED stereo camera: the left view provides the blurry input, while the right view serves as the sharp ground truth. Details of the blur synthesis process are described in \cite{zhou2019davanet}. For each scene, we extract 24 frames, and obtain camera extrinsics using COLMAP \cite{schonberger2016structure}. BARD-GS dataset are specifically collected for dynamic deblurring task. We further subselect frames that exhibit the most representative motions within each scene. 

\vspace{0.6em}
\noindent\textbf{Training Configurations.} Our model is trained on the blurry videos, then evaluated on the sharp counterparts. Either one of the rendered subframe (i.e start, middle, or end) is selected as the deblurred scene and subsequently compared with the sharp ground truth to perform the evaluation. Our method is trained for 200 epoch. All experiments were performed on Nvidia RTX 4080 32GB GPU.

\vspace{0.6em}
\noindent\textbf{Metrics.} We perform thorough quantitative comparisons by measuring the synthesized novel views with established metrics: Peak Signal-to-Noise Ratio (PSNR) \cite{huynh2008scope}, Structural Similarity Index Measure (SSIM) \cite{wang2004image}, and Learned Perceptual Image Patch Similarity (LPIPS) \cite{zhang2018unreasonable}, which together capture both reconstruction fidelity and perceptual quality.

\subsection{Comparison with State-of-the-Art Methods}

\begin{table}[t]
\centering
\footnotesize
\setlength{\tabcolsep}{3pt}
\renewcommand{\arraystretch}{1.05}
\caption{Deblurring results across three regions on the Stereo Blur dataset}
\resizebox{\linewidth}{!}{%
\begin{tabular}{l|c|c|c}
\toprule
\textbf{Methods} & \textbf{Dynamic region} & \textbf{Static region} & \textbf{Edge region} \\
\cmidrule(lr){2-2}\cmidrule(lr){3-3}\cmidrule(lr){4-4}
& \textbf{PSNR$\uparrow$/SSIM$\uparrow$/LPIPS$\downarrow$}
& \textbf{PSNR$\uparrow$/SSIM$\uparrow$/LPIPS$\downarrow$}
& \textbf{PSNR$\uparrow$/SSIM$\uparrow$/LPIPS$\downarrow$} \\
\midrule
Deblur4DGS & 28.283 / 0.817 / 0.125 & 29.672 / 0.716 / 0.182 & 24.844 / 0.869 / 0.135 \\
Ours       & \textbf{29.315} / \textbf{0.955} / \textbf{0.056} & \textbf{30.531} / \textbf{0.878} / \textbf{0.158} & \textbf{26.511} / \textbf{0.988} / \textbf{0.021} \\
\bottomrule
\end{tabular}}

\label{tab:deblur_regions}
\end{table}

\begin{figure}[t]
    \centering
    \includegraphics[width=0.8\linewidth, height=0.3\textheight, keepaspectratio]{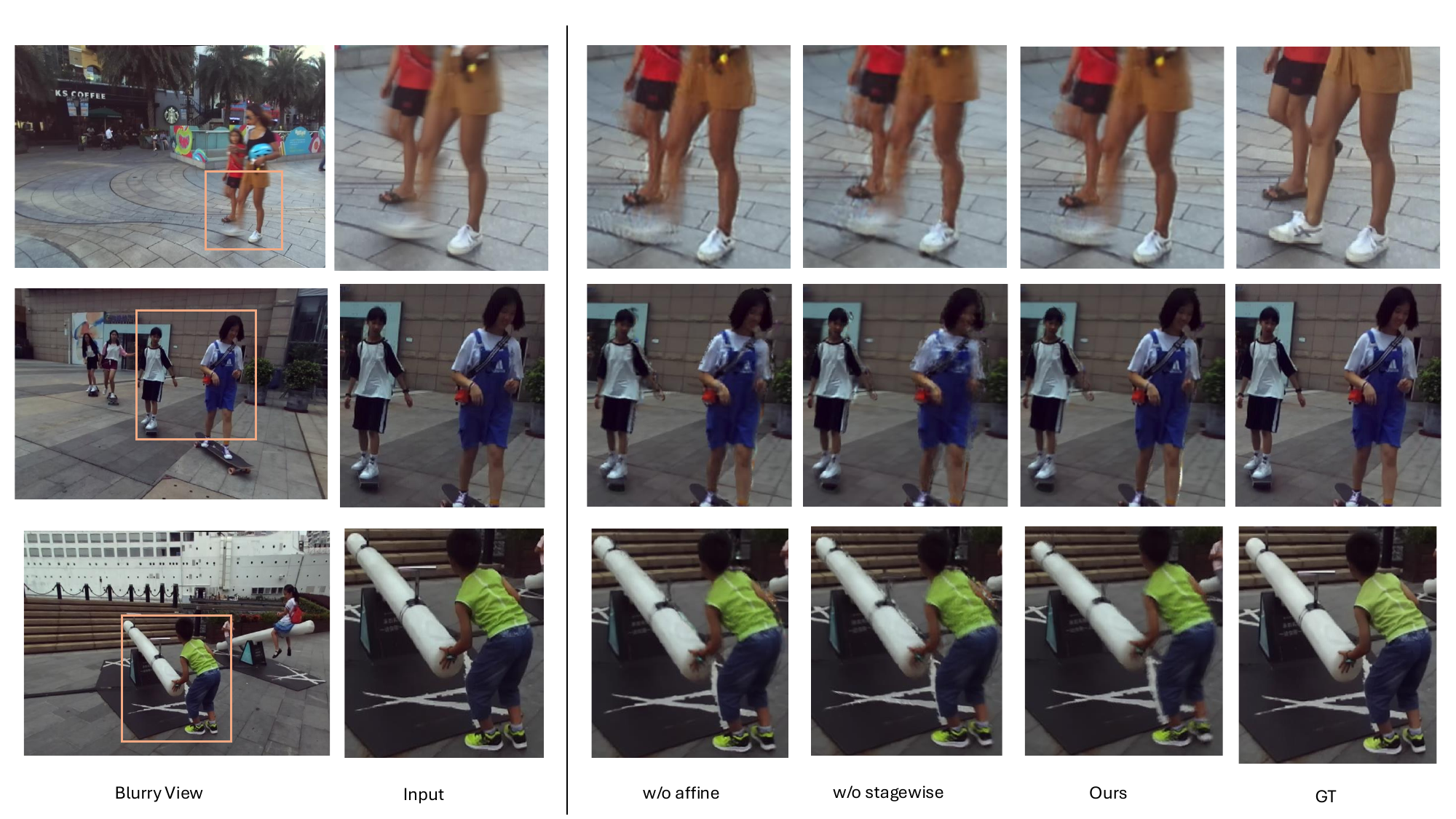} 
    \caption{\textbf{Visual Comparison of Ablation Studies} on a) Affine warp representation b) Stagewise optimization strategy}
    \label{fig:Abl-Jitter}
\end{figure}

\begin{figure}[t]
    \centering
    \includegraphics[width=0.7\linewidth, height=0.2\textheight, keepaspectratio]{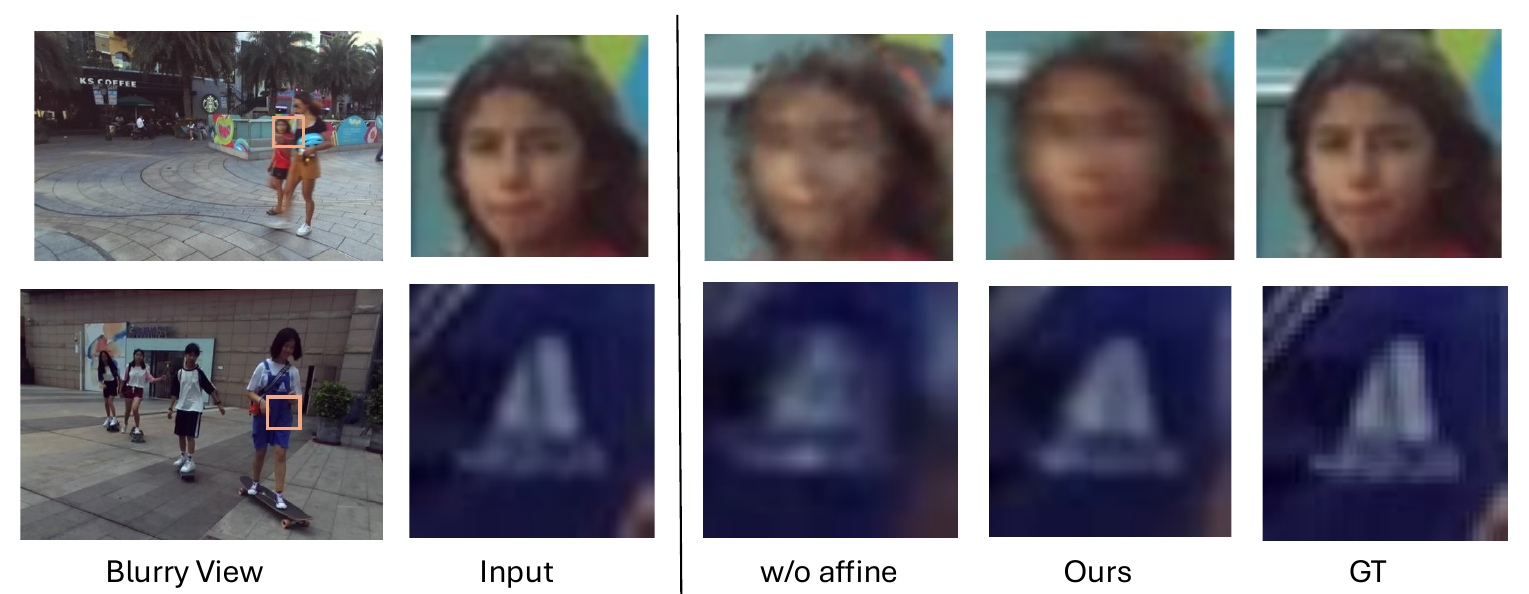} 
    \caption{\textbf{Ablation Study}: Effect of affine transformations on geometric modeling.}
    \label{fig:Abl-Affine}
\end{figure}

\textbf{Quantitative Results.}
The quantitative results on the Stereo Blur \cite{sun2024dyblurf} and BARD-GS dataset \cite{lu2025bard} are summarized in Table \ref{tab:avg_metrics}. We compare our pipeline among DyBluRF \cite{sun2024dyblurf}, Shape-of-Motion \cite{wang2024shape}, 4DGS \cite{wu20244d}, and Deblur4DGS \cite{wu2024deblur4dgs}. Our method attains the best PSNR and lowest LPIPS among all baselines, and a high SSIM that is close to the top score. Among the most comparable baselines, Deblur4DGS \cite{wu2024deblur4dgs} shares a Gaussian-splatting backbone and is designed specifically for motion-blurred novel-view synthesis. Against it, we improve PSNR by 1.12\,dB, SSIM by 0.166, and reduce LPIPS by 0.058 ($\approx$35\% relative). These gains indicate lower reconstruction error, higher structural fidelity, and substantially better perceptual quality, corresponding to sharper textures, cleaner boundaries, and fewer blur artifacts in the renders. 
In addition, Table \ref{tab:deblur_regions} reports the metrics on three regions—dynamic (FG), static (BG), and edges—to pinpoint where deblurring quality changes. The edge region is a boundary band around moving objects, where pose and motion errors couple most and artifacts tend to concentrate. Compared with Deblur4DGS (the closest 4DGS deblurring baseline), our method improves all three regions, with the largest gains on edges. As illustrated in the supplementary, our method achieves more significant improvements on scenes with challenging blurs, such as street, skating, and women. This further validates the superiority of our method.

\noindent\textbf{Visual Results.} We further compare the visual results produced by different methods in Fig. \ref{fig:sota}. Qualitatively, our reconstructions are sharper across both static backgrounds and moving subjects. In motion‑dominant regions where camera and object trajectories interact, our method produces crisper edges that align closely with ground‑truth boundaries, with far fewer jittered Gaussians along contours. Artifacts such as ghosts and background bleed‑through are largely suppressed. As further shown in the supplementary material, our results maintain continuous, uniform blur trails as objects move, reducing frame‑to‑frame flicker and the patchy over‑blending or under‑blending artifacts in the baseline results.

\subsection{Ablation Study}

The ablation study is conducted to investigate the effectiveness of the affine transformations and the optimiation strategy. The quantitative results are shown in Table \ref{tab:affine_stage_ours}.

\subsubsection{Affine Transformations}
Without affine transformations, subframe camera poses are decoupled from 3D Gaussians such that a notable performance drop is observed in Table \ref{tab:affine_stage_ours}. First, the quality of edge regions declines significantly. As shown in Fig. \ref{fig:Abl-Jitter}, jitter artifacts are obvious in boundary regions across all scenes. Second, the interior geometry deteriorates, with object bodies losing high‑frequency features and structural details. As shown in Fig. \ref{fig:Abl-Affine}, incorporating the affine transformations substantially improves geometric fidelity in regions with rich structure. For instance, facial features and clothing wrinkles are rendered much more clearly.

\begin{table}[t]
    \centering
    \caption{\textbf{Ablation studies} on affine warp and stagewise scene-pose optimization. Per-scene results are provided in the supplementary material.}

    {\small
    \setlength{\tabcolsep}{1.5pt}  
    \begin{tabular}{lcccccc}
        \hline
        & \multicolumn{3}{c}{\textbf{Stereo Dataset}}
        & \multicolumn{3}{c}{\textbf{BARD-GS Dataset}} \\
        & PSNR $\uparrow$ & SSIM $\uparrow$ & LPIPS $\downarrow$
        & PSNR $\uparrow$ & SSIM $\uparrow$ & LPIPS $\downarrow$ \\
        \hline
        w/o Affine    
            & 29.306 & 0.893 & 0.123
            & 18.772 & 0.734 & 0.528 \\
        w/o Stagewise 
            & 28.778 & 0.879 & 0.168
            & 18.421 & 0.677 & 0.531 \\
        \textbf{Ours} 
            & \textbf{30.143} & \textbf{0.911} & \textbf{0.107}
            & \textbf{19.502} & \textbf{0.731} & \textbf{0.550} \\
        \hline
    \end{tabular}
    }

    \label{tab:affine_stage_ours}
\end{table}

\begin{table}[t]
\centering
\caption{Ablations on number of latent frame and reference-frame evaluation. \colorbox{red!20}{Red} and \colorbox{yellow!20}{Yellow} indicate best and second best.}
\begin{subtable}[t]{0.49\linewidth}
\centering
\footnotesize
\setlength{\tabcolsep}{3pt}
\renewcommand{\arraystretch}{1.05}
\resizebox{0.92\linewidth}{!}{%
\begin{tabular}{lcccc}
\toprule
 & \textbf{PSNR}$\uparrow$ & \textbf{SSIM}$\uparrow$ & \textbf{LPIPS}$\downarrow$ & \textbf{Time} \\
\midrule
$N$=3  & 28.245 & 0.788  & 0.164  & 2.2h \\
$N$=5  & 28.998 & 0.847  & 0.142  & 3.4h \\
$N$=7  & \cellcolor{red!20}30.143  & \cellcolor{red!20}0.9118 & \cellcolor{red!20}0.1073 & \cellcolor{red!20}4.2h \\
$N$=9  & \cellcolor{yellow!20}30.098 & \cellcolor{yellow!20}0.9097 & \cellcolor{yellow!20}0.1123 & \cellcolor{yellow!20}5.1h \\
$N$=11 & 29.755 & 0.9028 & 0.1202 & 6.3h \\
$N$=13 & 29.945 & 0.9067 & 0.1152 & 6.9h \\
\bottomrule
\end{tabular}}
\caption{Effect of frame count.}
\label{tab:nframe}
\end{subtable}
\hfill
\begin{subtable}[t]{0.49\linewidth}
\centering
\footnotesize
\setlength{\tabcolsep}{3pt}
\renewcommand{\arraystretch}{1.05}
\resizebox{\linewidth}{!}{%
\begin{tabular}{@{}l|ccc|ccc@{}}
\toprule
\makecell{\textbf{Reference}\\\textbf{Frame}} &
\multicolumn{3}{c|}{\textbf{Stereo Blur}} &
\multicolumn{3}{c}{\textbf{BARD}} \\
\cmidrule(lr){2-4}\cmidrule(lr){5-7}
 & P$\uparrow$ & S$\uparrow$ & L$\downarrow$ & P$\uparrow$ & S$\uparrow$ & L$\downarrow$ \\
\midrule
Middle        & 29.17 & 0.867 & 0.134 & 18.33 & 0.680 & 0.624 \\
Sharpest (FG) & 29.49 & 0.874 & 0.114 & 19.19 & 0.725 & 0.596 \\
Sharpest      & 29.72 & 0.898 & 0.111 & 19.08 & 0.722 & 0.601 \\
\midrule
Ours          & \textbf{30.14} & \textbf{0.911} & \textbf{0.107} &
               \textbf{19.50} & \textbf{0.731} & \textbf{0.550} \\
\bottomrule
\end{tabular}}
\caption{Reference pose comparison.}
\label{tab:refpose}
\end{subtable}

\label{tab:two_tables_side_by_side}
\end{table}

\subsubsection{Optimization Strategy}
\begin{figure}[t]
    \centering
    \includegraphics[width=0.8\linewidth, height=0.5\textheight, keepaspectratio]{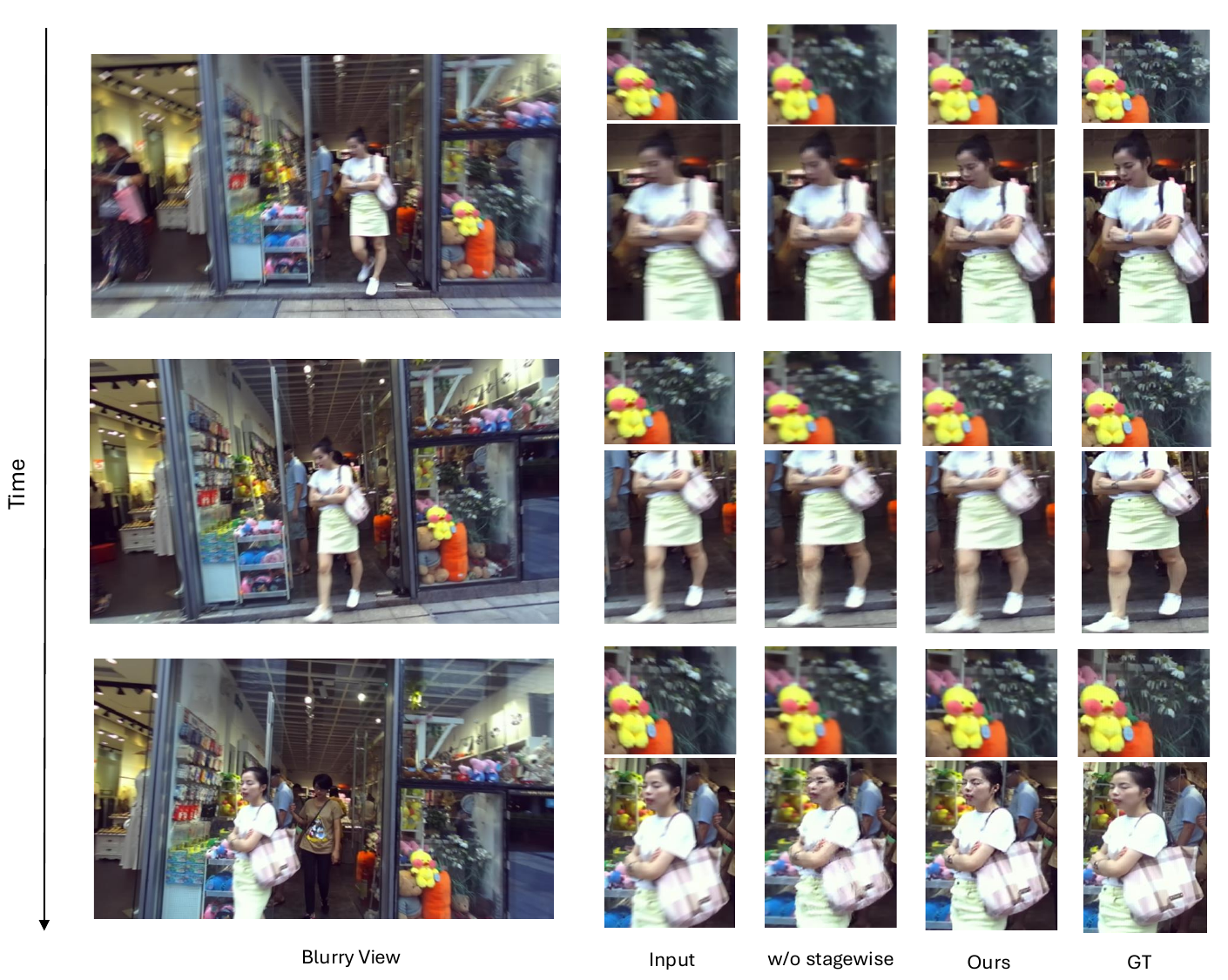} 
    \caption{\textbf{Ablation Studies}: Impact of our optimization strategy under large camera motion.}
    \label{fig:Abl-Stagewise}
\end{figure}

The stagewise training strategy reduces jitter and semi‑transparent Gaussians along object boundaries, as shown in Fig.~\ref{fig:Abl-Jitter}. Its benefit is most pronounced in scenes with noticeable camera motion.  In Fig.~\ref{fig:Abl-Stagewise} the camera undergoes large rotations and translations. By reconstructing the scene first with poses fixed, the subsequent pose-only phase estimates motion against an informative scene prior, which provides a strong geometric anchors for recovering the camera trajectory. This yields more accurate dynamic trajectories and markedly sharper static regions. The static portions of this scene are densely detailed, with intricate items that blur easily when the camera shakes, which makes deblurring difficult. With our training strategy, clean textures and crisp edges can be well reconstructed such that both foreground and background regions are rendered with noticeably higher quality.

\subsubsection{Latent Sharp Frame $N$}

We also conduct experiments on $N$ in Stereo Blur dataset, to investigate the number of latent sharp frames within an exposure, aiming to balance reconstruction quality and training time. As stated in Sec 3.1, $N$ is the count of latent sharp frames whose temporal average reproduces the observed blurry image. We sweep $N = 7$ to $13$. Table~\ref{tab:nframe} shows clear gains up to $N = 7$ and only marginal improvement beyond, while computational cost grows roughly linearly. We therefore set $N = 7$ for our final configuration.

\subsubsection{Reference pose sensitivity} 

We compare different reference pose selections in Table \ref{tab:refpose}, including the middle subframe, the sharpest subframe in the foreground region, and the sharpest subframe over the entire image. In our method, we select as reference the subframe that contains the largest number of visible foreground Gaussians. Empirically, this choice yields consistently strong performance across datasets. While selecting the sharpest foreground subframe produces comparable results, the performance differences are generally small. In our experiments, the sharpest subframe often lies close to the subframe with maximal foreground visibility, suggesting that these choices lead to similar optimization behavior in this setting.

\section{Conclusion}
We proposed \textbf{UPGS}, a unified, pose-aware Gaussian-splatting framework for dynamic deblurring that optimizes camera poses alongside 3D Gaussian primitives and simulates motion blur in the geometry domain with SE(3) transforms. The pipeline delivers higher reconstruction fidelity and more accurate poses than prior dynamic deblurring methods, with visible reductions in boundary jitter and blur artifacts. Experiments on Stereo Blur and challenging real footage confirm consistent gains over dynamic deblurring baselines. UPGS offers a practical path to robust 4D reconstruction from blurry monocular input.


%
%
\bibliographystyle{splncs04}
\bibliography{main}


\clearpage
\appendix
\section*{Supplementary Material}
\setcounter{section}{0}

\section{Unified Pose-aware Formulation}

Given a Gaussian set $\mathbf{G}$ and two camera poses $P^{(m)}$ and $P^{(n)}$, where each extrinsic is defined as:
\[
    P^{(\cdot)}
    =
    \begin{bmatrix}
        R^{(\cdot)} & t^{(\cdot)} \\
        0           & 1
    \end{bmatrix},
    \quad
    R^{(\cdot)} \in \mathbb{R}^{3\times3},\;
    t^{(\cdot)} \in \mathbb{R}^{3}.
\]
Our objective is to mimic the change of perspective from $P^{(m)}$ to $P^{(n)}$ by conducting an equivalent affine transformation on Gaussian under a fixed perspective (i.e., $P^{(m)}$). Mathematically, our objective is to learn an affine transformation to satisfy the following equation:
\[
    I\bigl(\mathcal{G},P^{(n)}\bigr)
    \;=\;
    I\bigl(T(\mathcal{G}),P^{(m)}\bigr),
\]
where $I(\mathcal{G},P^{(n)})$ denotes the image rendered from $\mathcal{G}$ under the camera pose $P^{(n)}$. Given camera intrinsics K, camera extrinsics P, and gaussian model G, we can then obtain:
\[
    \begin{aligned}
        T(\mathbf{G})
         & = \bigl(P^{(m)}\bigr)^{-1}\,P^{(n)}                                  \\[4pt]
         & = \begin{bmatrix}
                 R^{(m)\,\top} & -\,R^{(m)\,\top}t^{(m)} \\
                 0             & 1
             \end{bmatrix}
        \begin{bmatrix}
            R^{(n)} & t^{(n)} \\
            0       & 1
        \end{bmatrix}                                                       \\[4pt]
         & = \begin{bmatrix}
                 R^{(m)\,\top}R^{(n)} & R^{(m)\,\top}t^{(n)} - R^{(m)\,\top}t^{(m)} \\
                 0                    & 1
             \end{bmatrix}.
    \end{aligned}
\]

\begin{figure*}[t]
  \centering
  \includegraphics[width=1\linewidth]{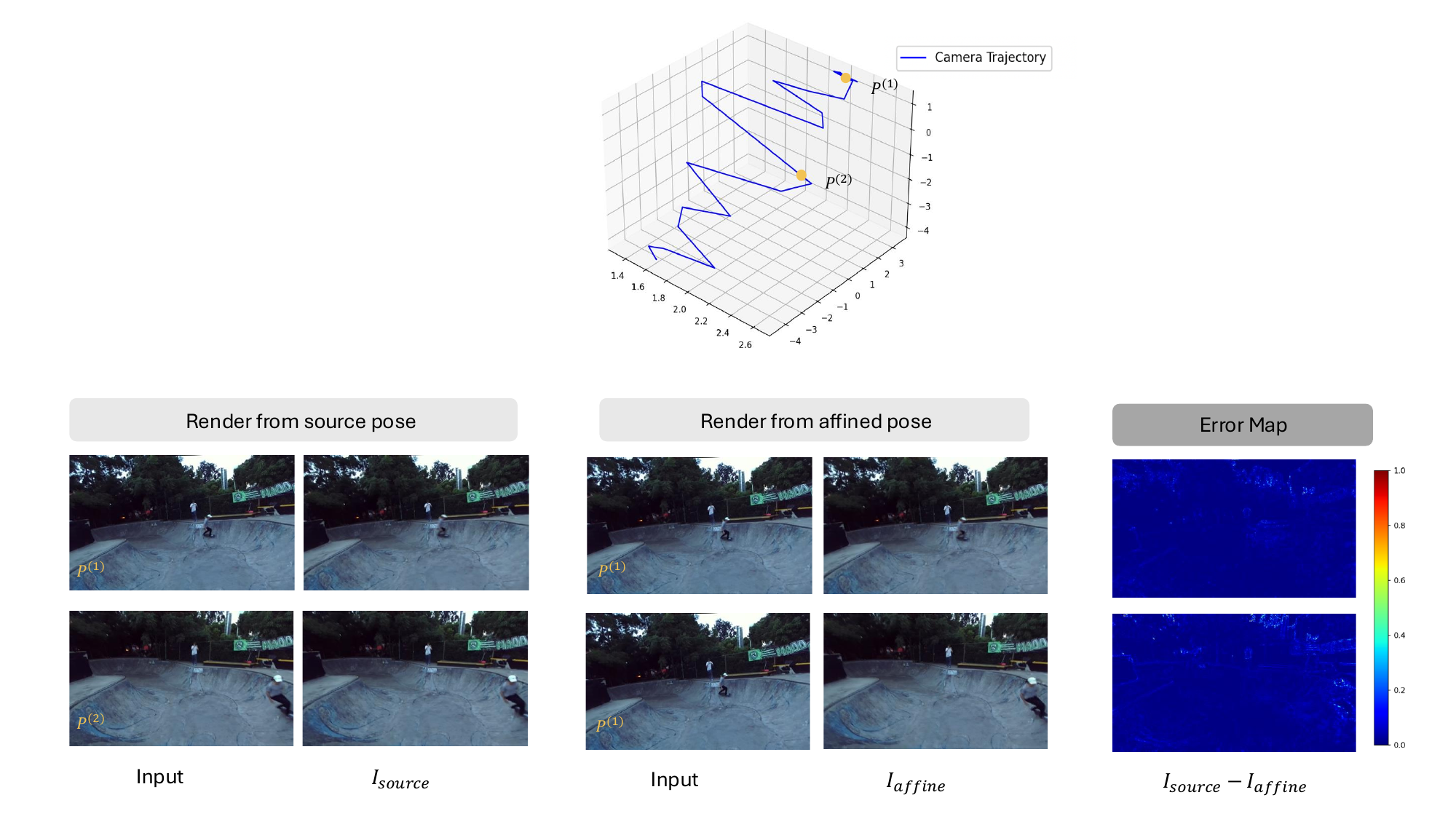} 
  \caption{\textbf{Validation of the unified pose-aware formulation.} Rendering the transformed Gaussians from a fixed reference pose produces images that are nearly identical to those rendered directly from the source poses, and the error maps confirm only negligible differences.}
  \label{fig:pose_comp}
\end{figure*}

\noindent Therefore, we have:
\[
    \begin{aligned}
        R_{T(\mathbf{G})}
         & = R^{(m)\,\top}R^{(n)},                        \\
        t_{T(\mathbf{G})}
         & = R^{(m)\,\top}t^{(n)} - R^{(m)\,\top}t^{(m)}.
    \end{aligned}
\]
This transformation can be applied directly to the Gaussians to mimic the perspective change from $P^{(m)}$ to $P^{(n)}$. In Fig.~\ref{fig:pose_comp}, we validate that rendering with our proposed formulation yields results nearly identical to those rasterized directly from the source pose. 

In the presence of motion blur, pose inaccuracies are prone to be absorbed into the scene geometry. Previous methods commonly use image-loss to implicitly drive camera pose optimization. As a result, these methods suffer inferior performance when blurry artifacts undermine the pose estimation accuracy. With our pose-aware formulation, we can introduce camera pose as learnable affine transformations that are compatible with 3DGS framework for end-to-end optimization, thereby producing superior performance.

\section{Gating Hyperparameter Specification}

To ensure reproducible gating behaviour across all experiments, we report the exact hyperparameter values used for the two adaptive gates as shown in Table~\ref{tab:gating_hyperparameters}.

\begin{table}[h]
\centering
\setlength{\tabcolsep}{10pt}
\begin{tabular}{lcc}
\hline
 & \textbf{Stereo Blur} & \textbf{BARD-GS} \\
\hline
$\epsilon_{1}$          & $2.4\times10^{-2}$   & $4.0\times10^{-2}$ \\
$\tau_a$                & $3.92\times10^{-5}$  & $1.92\times10^{-5}$ \\
$\epsilon_{2}$          & $7.0\times10^{-3}$   & $4.46\times10^{-5}$ \\
$\tau_{\mathrm{gain}}$  & $7.3\times10^{-4}$   & $3.0\times10^{-3}$ \\
\hline
\end{tabular}
\vspace{2mm}
\caption{Gating hyperparameters for the two adaptive gating criteria}
\label{tab:gating_hyperparameters}
\end{table}

\section{Additional Results and Ablations}
We record the full quantitative results for all scenes across both datasets, comparing our method against existing baselines. The metrics are summarized in Table \ref{tab:bards_gs_results}.
The ablation results on affine transformations and our training strategy across all six scenes are illustrated in Table \ref{tab:lpips_comparison}, which further demonstrate the effectiveness of our method designs.

\begin{table*}[t]
\centering
\footnotesize
\renewcommand{\arraystretch}{1.1}

\resizebox{\textwidth}{!}{%
\begin{tabular}{l|ccc|ccc|ccc}
\hline
& \multicolumn{3}{c|}{\textbf{Card}} 
& \multicolumn{3}{c|}{\textbf{Toycar}}
& \multicolumn{3}{c}{\textbf{Poster}} \\
\textbf{Method} 
& PSNR & SSIM & LPIPS
& PSNR & SSIM & LPIPS
& PSNR & SSIM & LPIPS \\
\hline
4DGS & 14.04 & 0.54 & 0.99 & 14.98 & 0.56 & 0.91 & 20.10 & 0.70 & 0.80 \\
Shape-of-motion & 15.41 & 0.74 & 0.50 & 16.25 & 0.75 & 0.43 & 20.86 & 0.75 & 0.40 \\
Deblur4DGS & 18.15 & 0.75 & 0.32 & 16.63 & 0.77 & 0.38 & 21.72 & 0.74 & 0.33 \\
Ours & 18.05 & 0.77 & 0.53 & 18.72 & 0.80 & 0.55 & 22.80 & 0.76 & 0.52 \\
\hline
\end{tabular}%
}

\vspace{1.0em}

\resizebox{\textwidth}{!}{%
\begin{tabular}{l|ccc|ccc|ccc}
\hline
& \multicolumn{3}{c|}{\textbf{Windmill}} 
& \multicolumn{3}{c|}{\textbf{Kitchen}}
& \multicolumn{3}{c}{\textbf{Shark-spin}} \\
\textbf{Method} 
& PSNR & SSIM & LPIPS
& PSNR & SSIM & LPIPS
& PSNR & SSIM & LPIPS \\
\hline
4DGS & 18.21 & 0.69 & 0.81 & 17.70 & 0.54 & 0.91 & 18.66 & 0.63 & 0.83 \\
Shape-of-motion & 17.75 & 0.76 & 0.41 & 18.59 & 0.62 & 0.45 & 18.65 & 0.68 & 0.42 \\
Deblur4DGS & 17.99 & 0.74 & 0.39 & 18.61 & 0.64 & 0.42 & 19.71 & 0.70 & 0.39 \\
Ours & 18.08 & 0.78 & 0.57 & 19.25 & 0.61 & 0.58 & 20.12 & 0.68 & 0.55 \\
\hline
\end{tabular}%
}
\vspace{1.0em}
\caption{Quantitative comparison across all six scenes (\textbf{Card, Toycar, Poster, Windmill, Kitchen, Shark-spin}) on the \textbf{BARD-GS} dataset. Metrics include PSNR, SSIM, and LPIPS.}
\label{tab:bards_gs_results}
\end{table*}

\section{Demo videos}
We also provide \texttt{demo.mov} for qualitative comparisons under real-world conditions. Compared with the baseline, our renderings demonstrate noticeably improved temporal consistency: object contours remain stable without jitter, textures are coherent over time with minimal flicker or shimmer, and motion-blur streaks appear smooth and continuous. These improvements are particularly evident in challenging regions where fast object motion overlaps with strong camera shake. The supplementary demo footage provides visual evidence of these effects, showcasing sharper, more stable novel views that complement the quantitative results presented in the paper.

\clearpage
\vspace{5.0em}
\begin{table*}[h]
    \centering
    \setlength{\tabcolsep}{6pt}
    \renewcommand{\arraystretch}{1.2}
    \begin{tabular}{lccccccc}
        \hline
        \textbf{PSNR} $\uparrow$ & \textbf{Street} & \textbf{Staking} & \textbf{Seesaw} & \textbf{Man} & \textbf{Women} & \textbf{Third} & \textbf{Avg} \\
        \hline
        Affine                   & 30.170          & 30.252           & 29.514          & 26.144       & 27.753         & 32.003         & 29.306       \\
        Stagewise                & 29.751          & 30.796           & 29.290          & 25.709       & 25.824         & 31.300         & 28.778       \\
        \textbf{Ours}            & 30.383          & 31.877           & 29.786          & 28.766       & 27.693         & 32.352         & 30.143       \\
        \hline
    \end{tabular}
    \vspace{1.0em}
    \label{tab:psnr_comparison}
\end{table*}

\vspace{-5.0em}

\begin{table*}[h]
    \centering
    \setlength{\tabcolsep}{6pt}
    \renewcommand{\arraystretch}{1.3}
    \begin{tabular}{lccccccc}
        \hline
        \textbf{SSIM} $\uparrow$ & \textbf{Street} & \textbf{Staking} & \textbf{Seesaw} & \textbf{Man} & \textbf{Women} & \textbf{Third} & \textbf{Avg} \\
        \hline
        Affine                   & 0.913           & 0.875            & 0.923           & 0.844        & 0.877          & 0.926          & 0.893        \\
        Stagewise                & 0.903           & 0.900            & 0.919           & 0.828        & 0.807          & 0.916          & 0.879        \\
        Ours                     & 0.920           & 0.915            & 0.929           & 0.906        & 0.869          & 0.932          & 0.911        \\
        \hline
    \end{tabular}
    \vspace{1.0em}
    \label{tab:ssim_comparison}
\end{table*}

\vspace{-5.0em}

\begin{table*}[h]
    \centering
    \setlength{\tabcolsep}{6pt}
    \renewcommand{\arraystretch}{1.3}
    \begin{tabular}{lccccccc}
        \hline
        \textbf{LPIPS} $\downarrow$ & \textbf{Street} & \textbf{Staking} & \textbf{Seesaw} & \textbf{Man} & \textbf{Women} & \textbf{Third} & \textbf{Avg} \\
        \hline
        Affine                      & 0.103           & 0.131            & 0.103           & 0.189        & 0.101          & 0.109          & 0.123        \\
        Stagewise                   & 0.133           & 0.158            & 0.107           & 0.215        & 0.244          & 0.151          & 0.168        \\
        Ours                        & 0.096           & 0.122            & 0.096           & 0.092        & 0.137          & 0.101          & 0.107        \\
        \hline
    \end{tabular}
    \vspace{1.0em}
    \caption{Quantitative comparison across all six scenes (\textbf{Street, Skating, Seesaw, Man, Women, Third}) on the \textbf{Stereo Blur} dataset. Metrics include PSNR, SSIM, and LPIPS.}
    \label{tab:lpips_comparison}
\end{table*}


\end{document}